\PassOptionsToPackage{table}{xcolor}
\documentclass[11pt]{article}

\usepackage[preprint]{acl}

\usepackage{times}
\usepackage{latexsym}

\usepackage[T1]{fontenc}

\usepackage{microtype}
\usepackage{hyperref}
\usepackage{url}
\usepackage{booktabs}
\usepackage{lineno}
\usepackage[pdftex]{graphicx}
\usepackage{array}
\usepackage{booktabs}
\usepackage{tabularx}
\usepackage{booktabs}
\usepackage{tabularx}
\usepackage{enumitem}
\usepackage{pdfpages}
\usepackage{fancyhdr}

\usepackage[utf8]{inputenc}

\usepackage{microtype}

\usepackage{inconsolata}

\usepackage{graphicx}

\usepackage{array}          
\usepackage{tabularx}       
\usepackage{siunitx}        
\usepackage{arydshln}

\makeatletter
\def\adl@drawiv#1#2#3{%
	\hskip.5\tabcolsep
	\xleaders#3{#2.5\@tempdimb #1{1}#2.5\@tempdimb}%
	#2\z@ plus1fil minus1fil\relax
	\hskip.5\tabcolsep}
\newcommand{\cdashlinelr}[1]{%
	\noalign{\vskip 2pt
		\global\let\@dashdrawstore\adl@draw
		\global\let\adl@draw\adl@drawiv}
	\cdashline{#1}[.4pt/2pt]
	\noalign{\global\let\adl@draw\@dashdrawstore
		\vskip 2pt}}
\makeatother

\newcolumntype{Y}{>{\raggedright\arraybackslash}X}

\definecolor{CustomBlue}{RGB}{57,83,191}
\definecolor{darkblue}{rgb}{0,0,0.5}
\hypersetup{
	colorlinks=true,
	citecolor=darkblue,
	linkcolor=darkblue,
	urlcolor=darkblue
}

\newcolumntype{Y}{>{\raggedright\arraybackslash}X}

\definecolor{CustomBlue}{RGB}{57,83,191}
\definecolor{darkblue}{rgb}{0,0,0.5}
\hypersetup{
	colorlinks=true,
	citecolor=darkblue,
	linkcolor=darkblue,
	urlcolor=darkblue
}

\title{Checklist Engineering Empowers Multilingual LLM Judges}

\author{Mohammad Ghiasvand Mohammadkhani \\
  Amirkabir University of Technology\\
  \texttt{mohammad.ghiasvand@aut.ac.ir} \\\And
  Hamid Beigy \\
  Sharif University of Technology \\
  \texttt{beigy@sharif.edu} \\}

\begin{document}
\maketitle
\begin{abstract}
	Automated text evaluation has long been a central issue in Natural Language Processing (NLP). Recently, the field has shifted toward using Large Language Models (LLMs) as evaluators—a trend known as the LLM-as-a-Judge paradigm. While promising and easily adaptable across tasks, this approach has seen limited exploration in multilingual contexts. Existing multilingual studies often rely on proprietary models or require extensive training data for fine-tuning, raising concerns about cost, time, and efficiency. In this paper, we propose \textbf{\textit{C}}hecklist \textbf{\textit{E}}ngineering based LLM-as-a-\textbf{\textit{Judge}} (\textbf{\textit{CE-Judge}}), a training-free framework that uses checklist intuition for multilingual evaluation with an open-source model. Experiments across multiple languages and three benchmark datasets, under both pointwise and pairwise settings, show that our method generally surpasses the baselines and performs on par with the GPT-4o model.\footnote{The code implementation is accessible at \url{https://github.com/mghiasvand1/CE-Judge}.}
\end{abstract}

\section{Introduction}

Evaluation is a fundamental task in Natural Language Processing (NLP) for measuring a model's performance on specific tasks. Automating this process offers significant benefits and has been a focus since the early stages of NLP research. Moreover, beyond creating evaluators proficient in English, it is crucial to develop their evaluation capabilities in parallel for other languages. Traditional evaluation metrics \citep{papineni2002bleu} have some drawbacks, such as the necessity of reference answers and a lack of interpretability, which has led to a paradigm shift toward developing Large Language Model (LLM) evaluators, referred to as LLM-as-a-Judge \citep{gu2025surveyllmasajudge,li2024generation}. These models are also capable of evaluating long-form LLM generations in either a pointwise or pairwise format—meaning grading a single response or selecting the better response out of two, respectively. Some advantages of this approach include high adaptability \citep{bavaresco2024llms} and interpretability, in contrast to traditional metrics, as well as low inference time and the fact that the evaluated LLM does not need to be active during evaluation (i.e., it does not need to generate additional responses), both in contrast to more complex LLM-based evaluation frameworks such as \citet{kim-etal-2025-llm-interviewer}.

Despite significant efforts to make LLMs multilingual \citep{qin2024multilingual}, extending LLM-as-a-Judge to multilingual configurations has received relatively little attention. Although current multilingual LLM judges \citep{pombal2025mprometheussuiteopenmultilingual, doddapaneni-etal-2025-cross} perform well, their main limitation is their reliance on proprietary models or the need for a large amount of real or synthetic data to fine-tune a capable evaluator, raising concerns regarding cost, time, and efficiency.

Meanwhile, checklists as interpretable evaluation tools \citep{doddapaneni2024finding, cook2024ticking} are gaining traction for their transparency and structure, although their application to multilingual evaluation remains relatively underexplored, and most of them also lack robust support for pairwise evaluation. For instance, \citep{wei2025rocketeval} suggests to handle the pairwise setting by selecting the higher-graded response based on independent pointwise scores, which fails to capture the nuanced comparative superiority between responses. In this work, we present CE-Judge, an LLM-as-a-Judge framework that builds and uses engineered checklists for evaluation. It supports multilingual evaluation and both pointwise and pairwise modes. Our pipeline follows a three-stage process: the first two stages aim to generate broad and dynamic checklist items, and the third applies them for judgment. Notably, by using a lightweight, open-source LLM without any fine-tuning, our method demonstrates strong performance across different evaluation scenarios.

\begin{figure*}[t]
	\centering
	\includegraphics[width=0.9\textwidth]{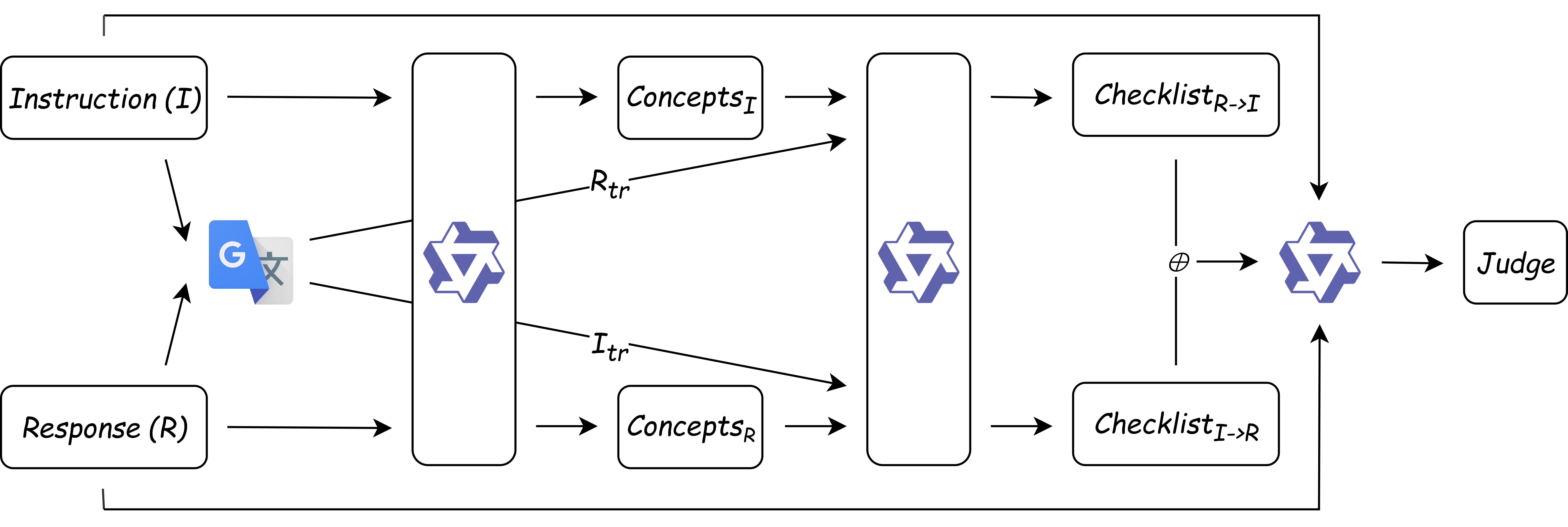}
	\caption{CE-Judge framework illustration.}
	\label{fig:framework}
\end{figure*}

\section{Related Works}

\subsection{LLM-as-a-Judge}
\label{rel:judge}
We begin with \citet{zheng2023judging}, which uses the generative capabilities of LLMs to act as evaluators. These can be grouped into two types: prompt-based and fine-tuned evaluators. For instance, for the former, \citet{li2025dna} adopts a "decompose and aggregate" strategy, identifying weighted evaluation aspects and combining their scores to assess candidate responses. For the latter, which has recently gained momentum, \citet{kim-etal-2024-prometheus} is a representative work that trains evaluators using large-scale synthetic data in both pointwise and pairwise setups, incorporating a weight-merging technique. In multilingual settings, consistency remains limited, as noted by \citet{fu2025reliable}. Among the few multilingual methods, \citet{doddapaneni-etal-2025-cross} translates between languages to anchor outputs in English for consistent scoring, while \citet{pombal2025mprometheussuiteopenmultilingual} follows \citet{kim-etal-2024-prometheus} to generate multilingual evaluation data and fine-tune models accordingly. \citet{chang2025exploringmultilingualnlgevaluation} investigates several aspects of multilingual LLM-based evaluators, including reference-free prompting, the effect of language resource availability, and the impact of fine-tuning. \citet{thellmann2024multilingualllmevaluationeuropean} creates various multilingual evaluation benchmarks while exploring the impact of translation and evaluating LLMs.

\subsection{Checklist-based Evaluators}
Several works have explored checklist-based evaluation. RocketEval \citep{wei2025rocketeval} generates binary checklist items, then reweights them to produce final scores. TICK \citep{cook2024ticking} uses instructions to generate checklists, which LLMs use for self-improvement. CheckEval \citep{lee2024checkeval} defines high-level criteria, then decomposes, diversifies, and filters them to form evaluation checklists. FBI \citep{doddapaneni2024finding} employs checklists for meta-evaluation to assess evaluator LLMs. Unlike these works, our framework introduces a novel architecture and, notably: (1) extends to multilingual settings; (2) supports pairwise evaluation beyond pointwise framing; and (3) uniquely incorporates broadness, descriptiveness, dynamism, and answer-mentioning in a unified manner.

\section{CE-Judge pipeline}

\begin{table*}[ht]
	\centering
	\small
	\setlength{\tabcolsep}{4pt}
	\renewcommand{\arraystretch}{1.4}
	\begin{tabularx}{0.9\textwidth}{%
			Y
			*{11}{S[table-format=1.2]}  
			S[table-format=2.2]          
		}
		\toprule
		\multicolumn{1}{l}{\textbf{Model}}
		& \multicolumn{11}{c}{\textbf{MMEval (Reasoning)}} 
		& \multicolumn{1}{c}{\textbf{Avg.}} \\
		\cmidrule(lr){2-12} \cmidrule(lr){13-13}
		& \textbf{en}
		& \textbf{de}
		& \textbf{fr}
		& \textbf{es}
		& \textbf{ru}
		& \textbf{zh}
		& \textbf{bn}
		& \textbf{ja}
		& \textbf{th}
		& \textbf{te}
		& \textbf{sw}
		& \\  
		\midrule
		
		\multicolumn{13}{l}{\textbf{Proprietary Models}} \\
		GPT-4o           & \textbf{0.79} & \underline{0.79} & \textbf{0.78} & \textbf{0.79} & \underline{0.76} & \textbf{0.78} & \textbf{0.84} & \underline{0.80} & \textbf{
			0.79} & \textbf{0.87} & \textbf{0.80} & \textbf{0.79} \\
		\cdashlinelr{1-13}
		
		\multicolumn{13}{l}{\textbf{Medium (7B parameters)}} \\
		Qwen2.5-7B-Instruct & 0.67 & 0.65 & 0.63 & 0.65 & 0.66 & 0.71 & 0.60 & 0.62 & 0.66 & 0.67 & 0.61 & 0.64 \\
		Hercule 7B & 0.50 & 0.56 & 0.55 & 0.55 & 0.53 & 0.57 & 0.57 & 0.54 & 0.52 & 0.54 & 0.51 & 0.54 \\
		M-Prometheus 7B   & 0.60 & 0.62 & 0.63 & 0.60 & 0.62 & 0.69 & 0.61 & 0.57 & 0.60 & 0.65 & 0.72 & 0.62 \\
		\cdashlinelr{1-13}
		
		\multicolumn{13}{l}{\textbf{Large (14B+ parameters)}} \\
		Prometheus 2 8x7B  & 0.54 & 0.65 & 0.58 & 0.58 & 0.58 & 0.64 & 0.57 & 0.56 & 0.60 & 0.60 & 0.63 & 0.59 \\
		M-Prometheus 14B  & 0.64 & 0.70 & 0.70 & 0.69 & 0.69 & 0.72 & 0.70 & 0.70 & 0.68 & 0.72 & 0.76 & 0.70 \\
		\cdashlinelr{1-13}
		
		\multicolumn{13}{l}{\textbf{Ours (7B parameters)}} \\
		\rowcolor{CustomBlue!10}
		CE-Judge         & \underline{0.77} & \textbf{0.81} & \underline{0.77} & \underline{0.72} & \textbf{0.78} & \underline{0.77} & \underline{0.75} & \textbf{0.84} & \underline{0.78} & \underline{0.76} & \underline{0.78} & ~~\underline{0.77} \\
		\bottomrule
	\end{tabularx}
	\caption{Accuracy on MMEval (Reasoning) broken down by language.}
	\label{tab:res}
\end{table*}

We present our training-free, efficient evaluation framework (Figure~\ref{fig:framework}), which consists of three steps. The pipeline first constructs an engineered checklist through level-by-level multilingual understanding and integration of input-output linkages to enable dynamism, followed by utilizing this checklist to enhance the decisions of the evaluator LLM. All LLM generations are asked to be in English to leverage its strong performance in the language \citep{mondshine-etal-2025-beyond}.

\subsection{Concepts generation}
Considering an instruction as input—replaced by the source text in the translation evaluation task—and a corresponding response, we pass each separately to the LLM along with the prompt in \ref{app:concept} for concept generation. This generation aims to produce an abstract-level text that represents the skeleton of the corresponding text.

\subsection{Checklist generation}
Next, we translate both the instruction and response texts into English. Using the translated response, the concepts generated from the instruction (from the previous step), and the prompt in \ref{app:check}, we generate a checklist following the “response to instruction” direction. Likewise, we use the translated instruction and the response’s concepts to generate a checklist for the “instruction to response” direction. This dual approach aims to blind each side once, broadening checklist coverage and enhancing awareness of both sides’ content, rather than relying on a standard checklist with limited, predefined criteria. In this step, we also avoid prejudgment and ask the model to generate more descriptive items, going beyond simple binary questions.

\subsection{Judgment}
The final step is judgment. First, the two checklists from the previous steps are concatenated into a unified checklist. It’s important to note that the entire process described so far is for pointwise evaluation. For pairwise evaluation, the process remains the same, except that the previous two steps are applied to two candidate responses instead of one. As a result, after concatenation, we obtain two checklists, one for each candidate. In this step, we provide the untranslated versions of the instruction and response(s)—to avoid translation bias—along with the checklist(s) and the prompt template in \ref{app:judge}. The LLM is then asked to answer a subset of key checklist items and generate evaluation feedback. Unlike prior works where checklist items are marked with ticks, crosses, or weighted scores, here the model exercises discretion in its judgments, and the final evaluation is left to the model’s decision.

\section{Experiments}

\begin{table*}[ht]
	\centering
	\small
	\setlength{\tabcolsep}{5pt}
	\renewcommand{\arraystretch}{1.4}
	\begin{tabularx}{0.71\textwidth}{%
			Y
			*{11}{S[table-format=1.2]}  
			S[table-format=2.2]          
		}
		\toprule
		\multicolumn{1}{l}{\textbf{Model}}
		& \multicolumn{7}{c}{\textbf{MMEval (Chat)}} 
		& \multicolumn{1}{c}{\textbf{Avg.}} \\
		\cmidrule(lr){2-8} \cmidrule(lr){9-9}
		& \textbf{en}
		& \textbf{de}
		& \textbf{fr}
		& \textbf{es}
		& \textbf{ca}
		& \textbf{ru}
		& \textbf{zh}
		& \\
		\midrule
		
		\multicolumn{9}{l}{\textbf{Proprietary Models}} \\
		GPT-4o           & \textbf{0.72} & 0.70 & \textbf{0.73} & 0.64 & 0.75 & \textbf{0.78} & 0.80 & \underline{0.73} \\
		\cdashlinelr{1-12}
		
		\multicolumn{9}{l}{\textbf{Medium (7B parameters)}} \\
		Qwen2.5-7B-Instruct      & \underline{0.69} & \textbf{0.75} & 0.71 & \textbf{0.78} & 0.72 & 0.66 & \underline{0.85} & 0.72 \\
		Hercule 7B    & 0.62 & 0.71 & 0.61 & 0.55 & 0.62 & 0.64 & 0.65 & 0.62 \\
		M-Prometheus 7B       & 0.68 & 0.65 & 0.66 & 0.59 & 0.62 & 0.56 & 0.58 & 0.62 \\
		\cdashlinelr{1-12}
		
		\multicolumn{9}{l}{\textbf{Large (14B+ parameters)}} \\
		Prometheus 2 8x7B      & 0.64 & 0.68 & \underline{0.72} & 0.65 & \underline{0.77} & 0.64 & 0.80 & 0.70 \\
		M-Prometheus 14B       & 0.61 & \underline{0.72} & 0.64 & 0.64 & 0.57 & 0.71 & 0.73 & 0.66 \\
		\cdashlinelr{1-12}
		
		\multicolumn{9}{l}{\textbf{Ours (7B parameters)}} \\
		\rowcolor{CustomBlue!10}
		CE-Judge         & \underline{0.69} & 0.60 & \textbf{0.73} & \underline{0.77} & \textbf{0.82} & \underline{0.77} & \textbf{0.87} & \textbf{0.75} \\
		\bottomrule
	\end{tabularx}
	\caption{Accuracy on MMEval (Chat) broken down by language.}
	\label{tab:chat}
\end{table*}

\subsection{Experimental setup}

In this work, we used the \texttt{Qwen2.5-7B-Instruct} model \citep{yang2024qwen2} as the backbone LLM, accessed freely via the \textit{Novita API}\footnote{\url{https://novita.ai/}}. The hyperparameters ``temperature'', ``top\_p'', and ``seed'' were set to 0, 1, and 42, respectively, to ensure reproducibility. For translation, we employed the free \textit{Google Translate API} available through the \texttt{deep-translator} Python package\footnote{\url{https://deep-translator.readthedocs.io/en/latest/README.html}}.

\subsection{Datasets}
Since our method is training-free, all datasets are used solely for testing. We evaluated our framework in both pointwise and pairwise settings. For pointwise evaluation, we used the student-annotated subset of the LitEval \citep{zhang2024good}, which contains source–target literary translations for four language pairs with human ratings from 1 to 7. For pairwise evaluation, we employed the reasoning and chat subsets of the MM-Eval dataset \citep{son2024mm}, covering 11 and 7 languages, respectively. Each input consists of a reasoning question or chat history, with the task being to choose the better of two candidate responses. The reason for utilizing the LitEval and MM-Eval datasets is that the former is one of the only multilingual pointwise evaluation datasets, and the latter is more robust than the well-known M-RewardBench multilingual benchmark \citep{gureja-etal-2025-rewardbench}.

\subsection{Evaluation Metrics}

To evaluate our CE-Judge framework in pointwise mode, we measured performance using Kendall’s Tau correlation coefficient \citep{kendall1938new}, which assesses agreement between our model’s rankings and human judgments. For the pairwise setting, we used accuracy—defined as the number of correct predictions over the total number of samples.

\begin{table*}[ht]
	\centering
	\small
	\setlength{\tabcolsep}{3pt}
	\renewcommand{\arraystretch}{1.4}
	\begin{tabularx}{0.59\textwidth}{%
			Y
			*{4}{S[table-format=1.2]}
			S[table-format=2.2]
		}
		\toprule
		\multicolumn{1}{l}{\textbf{Model}}
		& \multicolumn{4}{c}{\textbf{LitEval}}
		& \multicolumn{1}{c}{\textbf{Avg.}} \\
		\cmidrule(lr){2-5} \cmidrule(lr){6-6}
		& \textbf{de$\rightarrow$en}
		& \textbf{en$\rightarrow$de}
		& \textbf{en$\rightarrow$zh}
		& \textbf{de$\rightarrow$zh}
		&                                  \\
		\midrule
		
		\multicolumn{6}{l}{\textbf{Proprietary Models}} \\
		GPT-4o           & 0.26 & 0.48 & 0.41 & 0.40 & 0.38 \\
		\cdashlinelr{1-6}
		
		\multicolumn{6}{l}{\textbf{Medium (7B parameters)}} \\
		Qwen2.5-7B-Instruct & 0.12 & 0.32 & 0.17 & 0.07 & 0.17 \\
		Hercule 7B          & 0.26            & 0.33           & 0.38 & 0.42          & 0.34 \\
		M-Prometheus 7B     & 0.20           & \underline{0.53}          & 0.46 & \underline{0.54}         & ~~\underline{0.43} \\
		\cdashlinelr{1-6}
		
		\multicolumn{6}{l}{\textbf{Large (14B+ parameters)}} \\
		Prometheus 2 8x7B   & 0.24            & 0.36           & 0.25 & 0.40 & 0.31 \\
		M-Prometheus 14B    & \textbf{0.29}            & \textbf{0.57} & \underline{0.48} & \textbf{0.56} & ~~\textbf{0.47} \\
		\cdashlinelr{1-6}
		
		\multicolumn{6}{l}{\textbf{Ours (7B parameters)}} \\
		\rowcolor{CustomBlue!10}
		CE-Judge           & \underline{0.28} & 0.46 & \textbf{0.49} & 0.30 & 0.38 \\
		\bottomrule
	\end{tabularx}
	\caption{Kendall correlation on LitEval broken down by language pair.}
	\label{tab:liteval}
\end{table*}

\subsection{Baselines}
We compare our framework with three types of models. The first includes proprietary models like \texttt{GPT-4o}. The second is \texttt{Qwen2.5-7B-Instruct}, a strong multilingual open-source LLM that is instruction-tuned from a pretrained model without further fine-tuning. The third category consists of models explicitly trained as evaluators, such as \texttt{Prometheus 2} \citep{kim-etal-2024-prometheus}, \texttt{Hercule} \citep{doddapaneni-etal-2025-cross}, and \texttt{M-Prometheus} \citep{pombal2025mprometheussuiteopenmultilingual}, as discussed in Subsection~\ref{rel:judge}.

\section{Results}

We evaluate CE-Judge on three multilingual evaluation datasets—reasoning, chat, and literary translation—against proprietary and open-source baselines, including the fine-tuned M-Prometheus. In all three tables, languages are shown by their codes, and, more importantly, the results for the other models are taken from \citet{pombal2025mprometheussuiteopenmultilingual}.

\begin{itemize}
	\item In the reasoning evaluation task (Table~\ref{tab:res}), CE-Judge achieves an average accuracy of \textbf{0.77}, outperforming all open-source baselines in all languages, including large fine-tuned evaluators such as M-Prometheus 14B. Despite being training-free and based on a 7B-parameter model, it performs competitively with GPT-4o (which has an average accuracy of 0.79) and maintains strong performance across both high- and low-resource languages.
	
	\item In the chat evaluation (Table~\ref{tab:chat}), CE-Judge achieves an average accuracy of \textbf{0.75}, surpassing GPT-4o (with the average of 0.73) and significantly outperforming the M-Prometheus models across nearly all languages. This result highlights the robustness of our checklist-driven approach in conversational scenarios that require nuanced, context-aware judgment.
	
	\item In the literary translation evaluation (Table~\ref{tab:liteval}), which requires nuanced linguistic and stylistic understanding, CE-Judge achieves an average Kendall’s Tau correlation of \textbf{0.38}, significantly outperforming its backbone model, Qwen2.5-7B, and delivering performance comparable to GPT-4o. Although it slightly lags behind M-Prometheus 7B (average of 0.43)—which benefits from fine-tuning on supervised machine translation evaluation data—our training-free approach remains highly competitive.
\end{itemize}

\section{Conclusion}
In this work, we introduce CE-Judge, a novel and straightforward checklist-based framework for multilingual LLM-as-a-Judge that is training-free and built on an open-source model. By leveraging dynamic, broad, and flexible checklist items, CE-Judge supports both pointwise and pairwise evaluations across diverse languages. Experiments on multiple multilingual benchmarks show that CE-Judge not only generally outperforms open-source fine-tuned baselines but also performs on par with GPT-4o. These results highlight the promise of structured, dynamic evaluation techniques for improving the reliability and interpretability of LLM judgment, particularly in multilingual contexts, for more consistent performance.

\section*{Ethics Statement}
This study aims to advance multilingual evaluation using a training-free approach built on an open-source LLM, prioritizing accessibility and transparency. We leveraged publicly available datasets and APIs, with no collection of personal or sensitive data. All experiments are free from human involvement and pose no privacy or safety risks.

\section*{Limitations}

Despite its strong results and training-free design, our framework has several limitations to address in future work. First, crafting effective prompts for each of the three steps per task can be time-consuming, so developing an automatic, adaptable prompt generation module would be beneficial. Second, our method relies solely on LLM generation, which may suffer from misalignment between training objectives and robust text generation. Incorporating internal LLM representations, as shown by \citet{sheng2024repeval}, could capture more accurate implicit knowledge. Finally, our framework’s flexibility suggests potential extensions as a plug-and-play method or adaptations to other evaluation strategies, such as interview-based evaluation \citep{kim-etal-2025-llm-interviewer}.

\bibliography{custom}

\appendix

\section{Prompt templates}

\subsection{Concepts Generation Prompts}
\label{app:concept}
\begin{figure*}[h]
	\centering
	\includegraphics[page=1,width=0.8\textwidth]{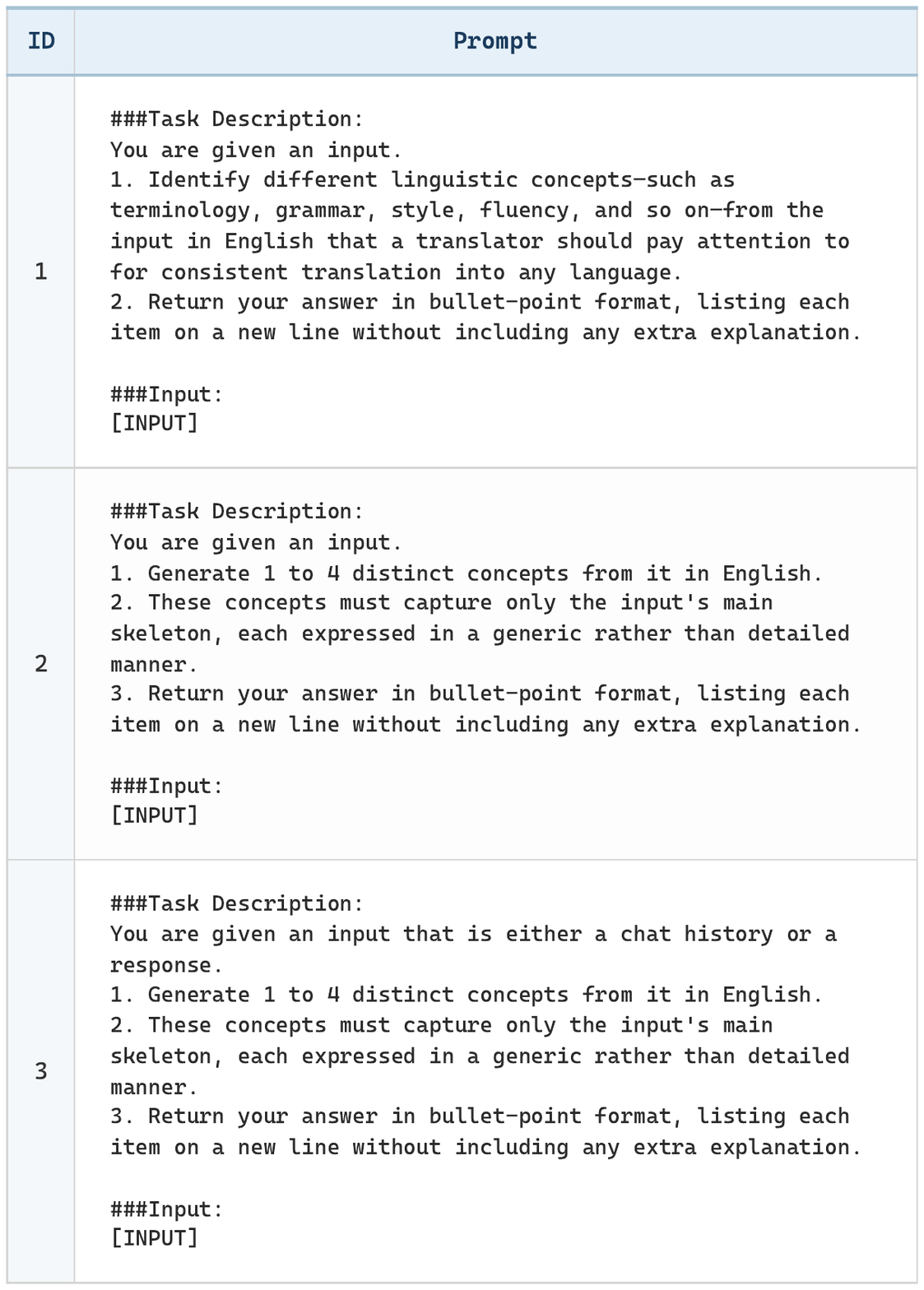}
	\caption{Concept generation prompts for LitEval and MM-Eval (Reasoning \& Chat) datasets.}
	\label{fig:concept}
\end{figure*}

The prompts for this step, across all three datasets, are shown in \ref{fig:concept}, and the “[INPUT]” placeholder must be replaced with the text from which we want to extract concepts, such as an instruction, response, etc.

\subsection{Checklist Generation Prompts}
\label{app:check}
\begin{figure*}[h]
	\centering
	\includegraphics[page=1,width=0.9\textwidth]{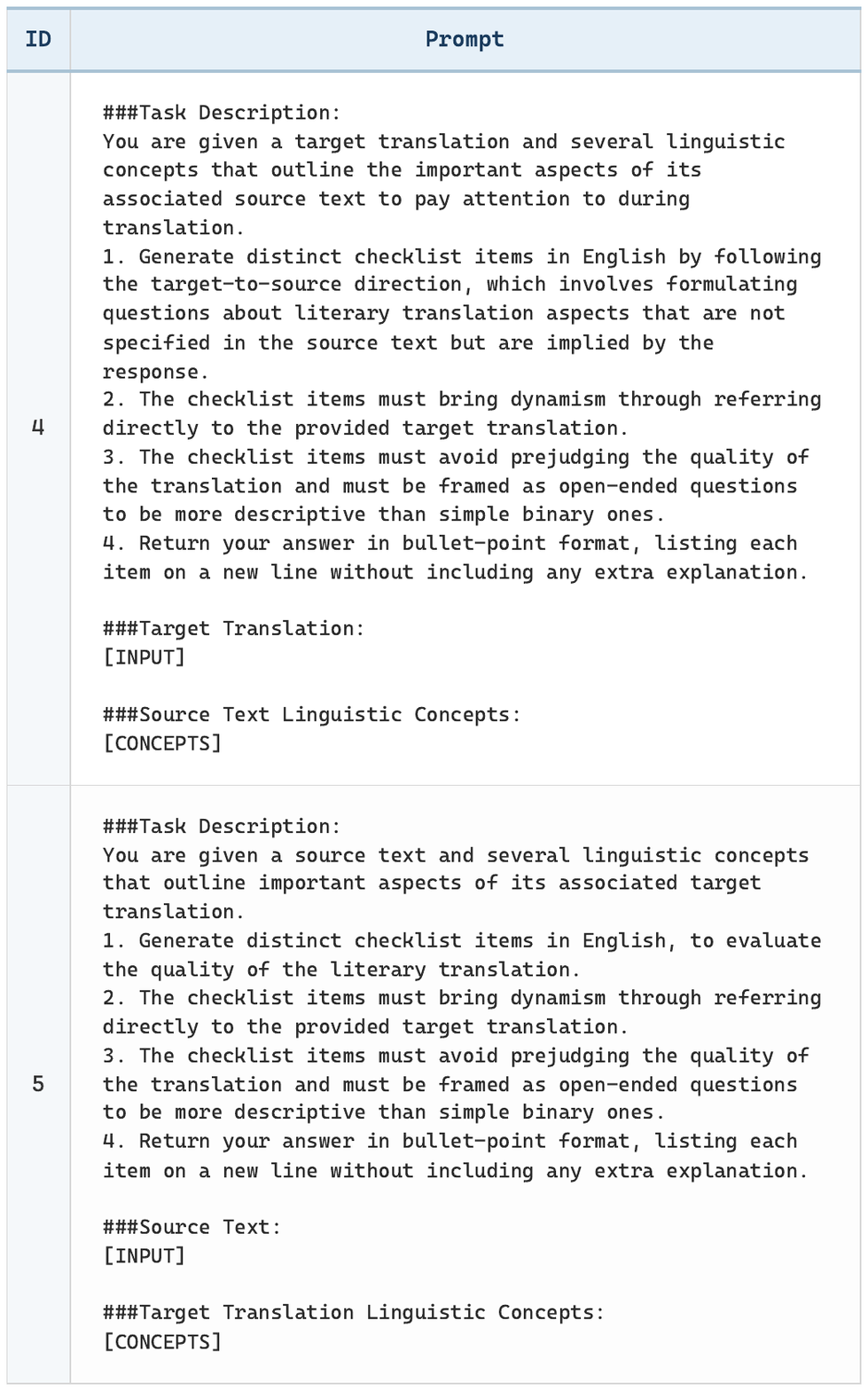}
	\caption{Checklist generation prompts for LitEval dataset.}
	\label{fig:checklit}
\end{figure*}

\begin{figure*}[h]
	\centering
	\includegraphics[page=1,width=0.9\textwidth]{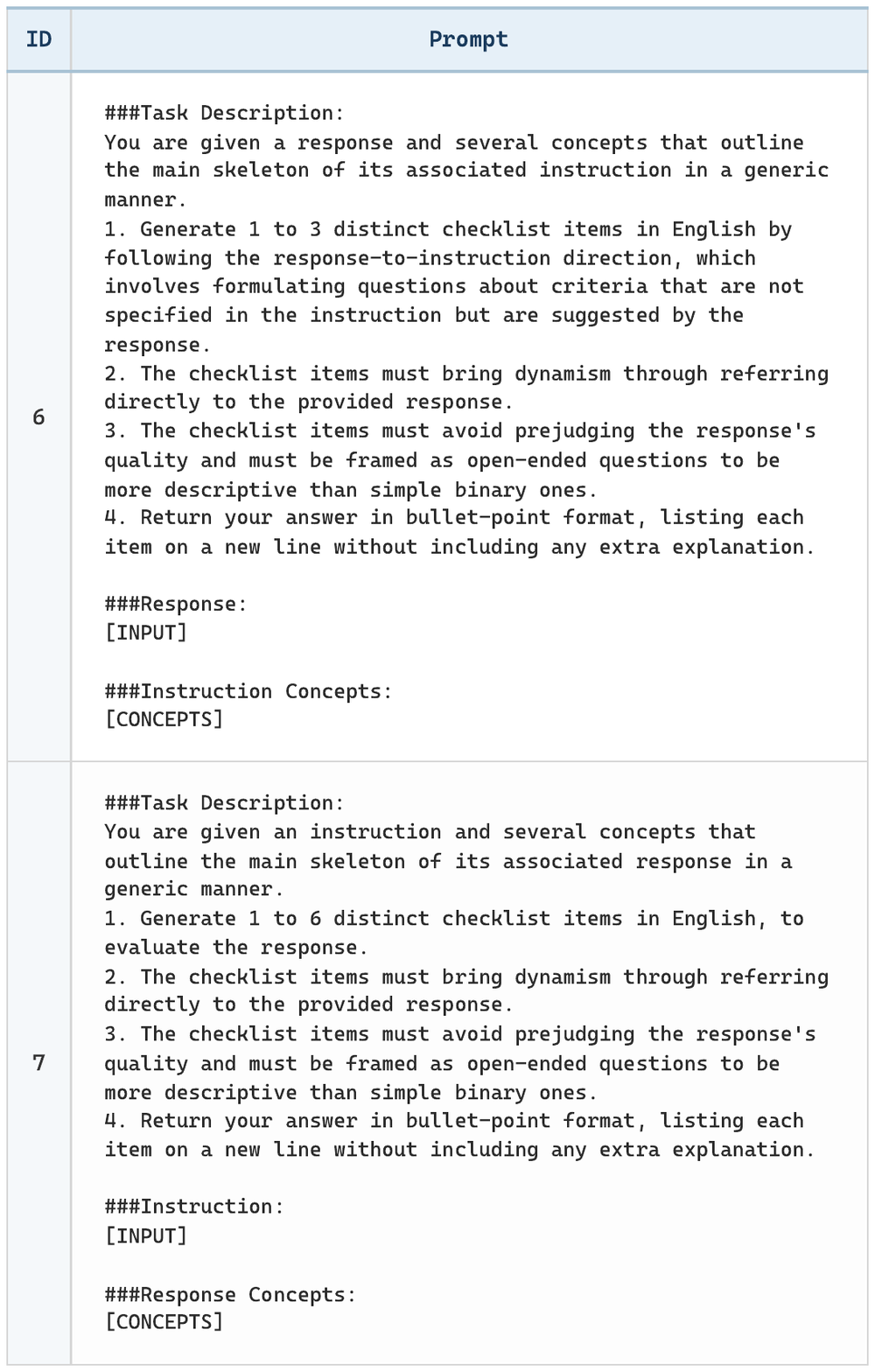}
	\caption{Checklist generation prompts for MM-Eval (Reasoning) dataset.}
	\label{fig:checkreason}
\end{figure*}

\begin{figure*}[h]
	\centering
	\includegraphics[page=1,width=0.9\textwidth]{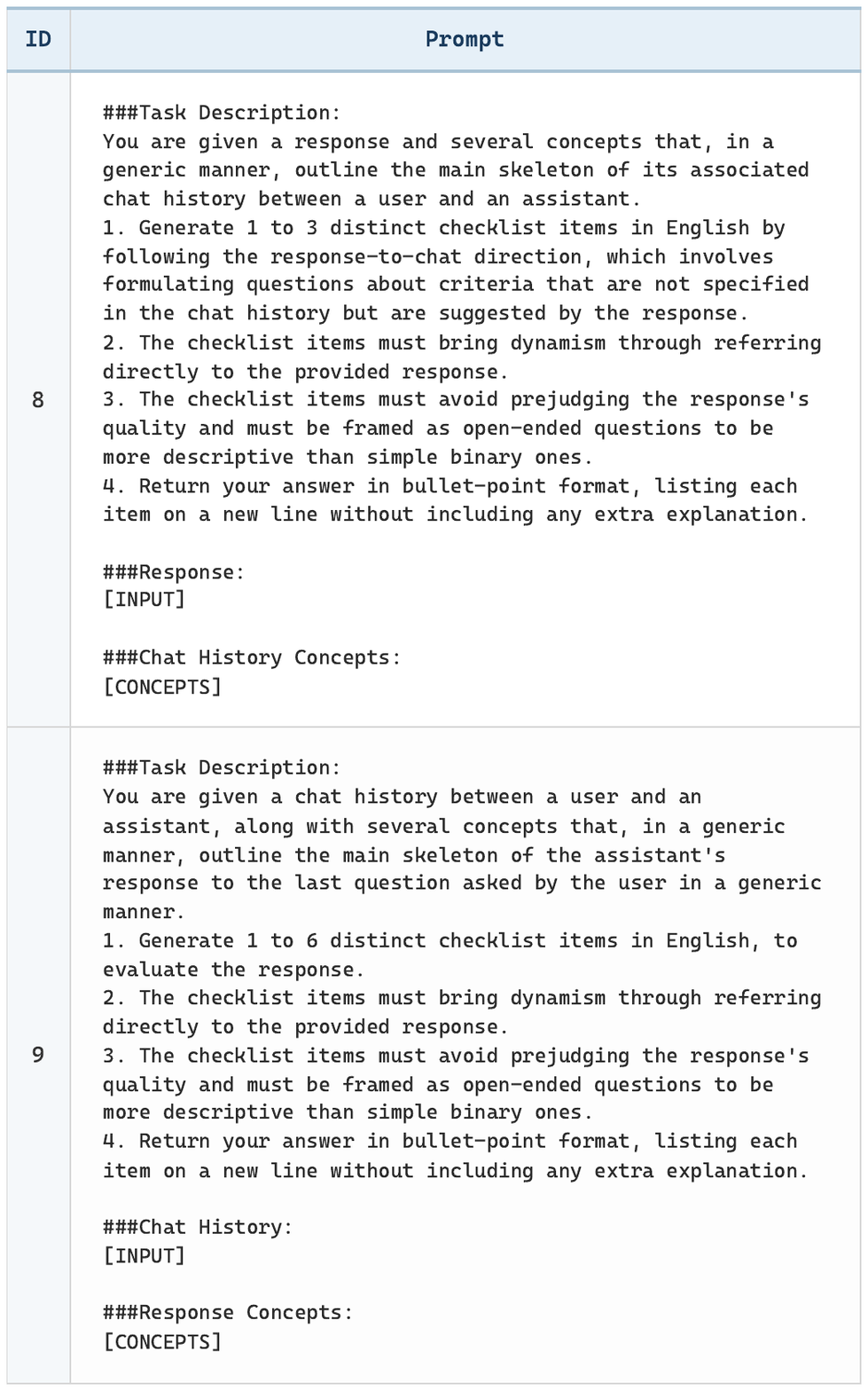}
	\caption{Checklist generation prompts for MM-Eval (Chat) dataset.}
	\label{fig:checkchat}
\end{figure*}

Figures \ref{fig:checklit}, \ref{fig:checkreason}, and \ref{fig:checkchat} show checklist generation prompts for Liteval, MM-Eval (Reasoning), and MM-Eval (Chat), respectively. Each figure consists of two prompts indicating the checklist creation direction. Note that the “[CONCEPTS]” placeholder must be replaced with the concepts generated in the previous step.

\subsection{Judgment Prompts}
\label{app:judge}
\begin{figure*}[h]
	\centering
	\vspace{-0.5em} 
	\includegraphics[page=1,width=0.8\textwidth]{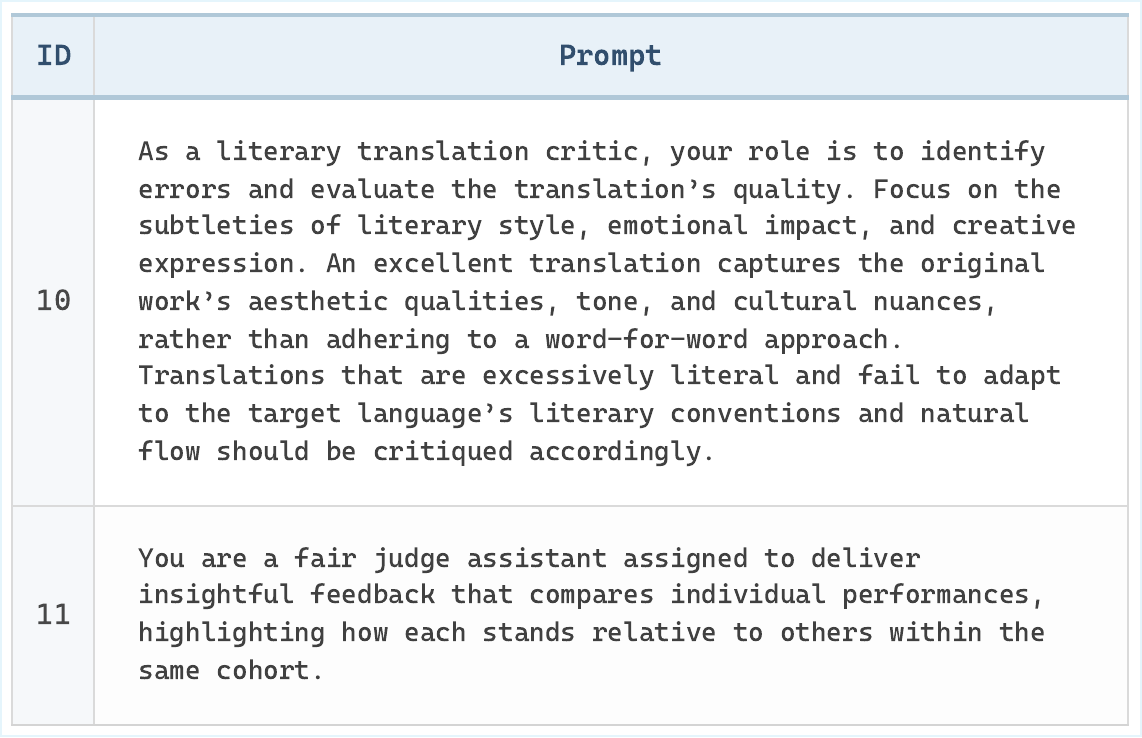}
	
	\caption{Judgment system prompts for LitEval and MM-Eval (Reasoning \& Chat) datasets.}
	\label{fig:system}
\end{figure*}

\begin{figure*}[h]
	\centering
	\includegraphics[page=1,width=0.9\textwidth]{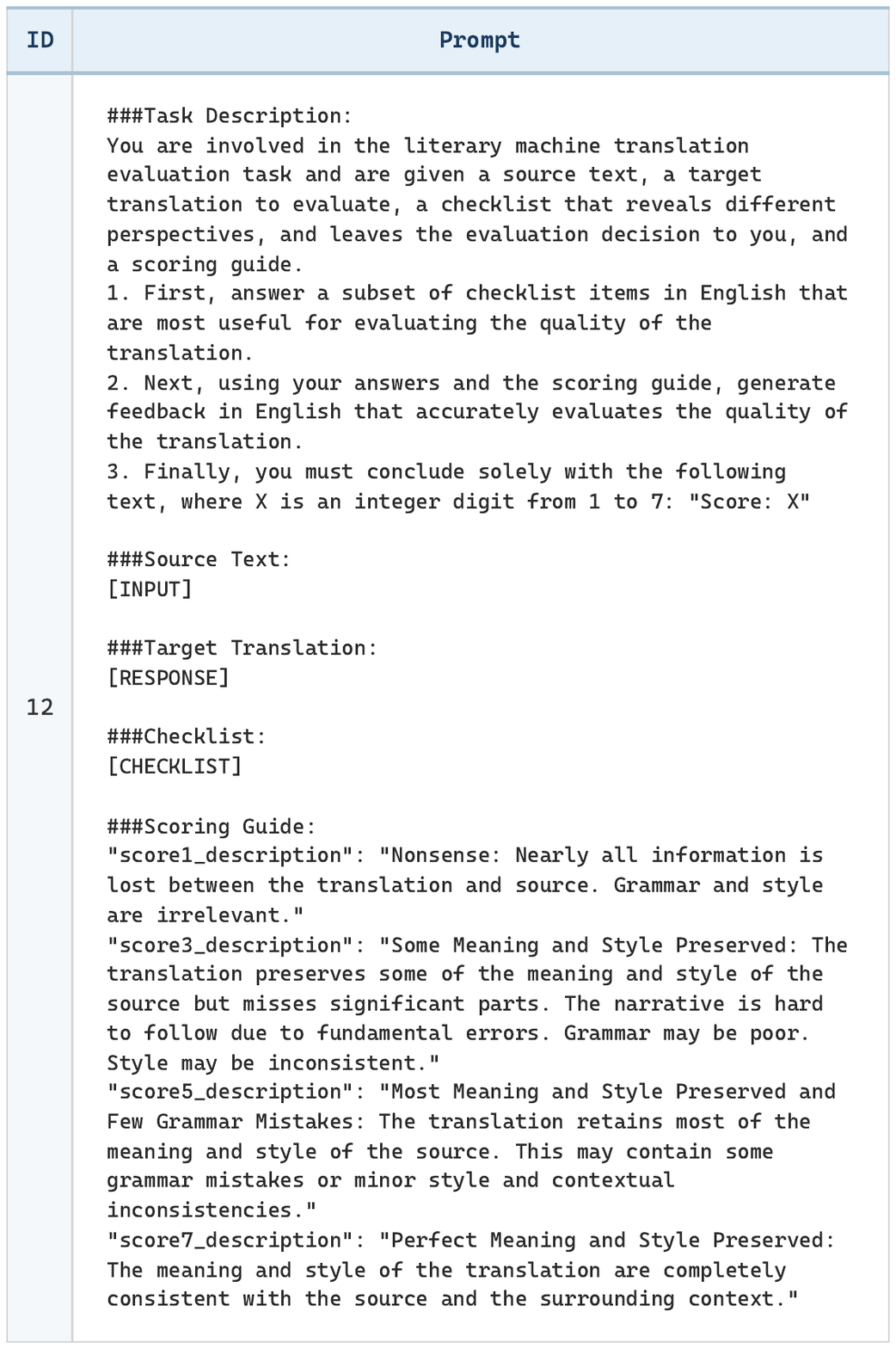}
	\caption{Judgment prompt for LitEval dataset.}
	\label{fig:judgelit}
\end{figure*}

\begin{figure*}[h]
	\centering
	\includegraphics[page=1,width=1\textwidth]{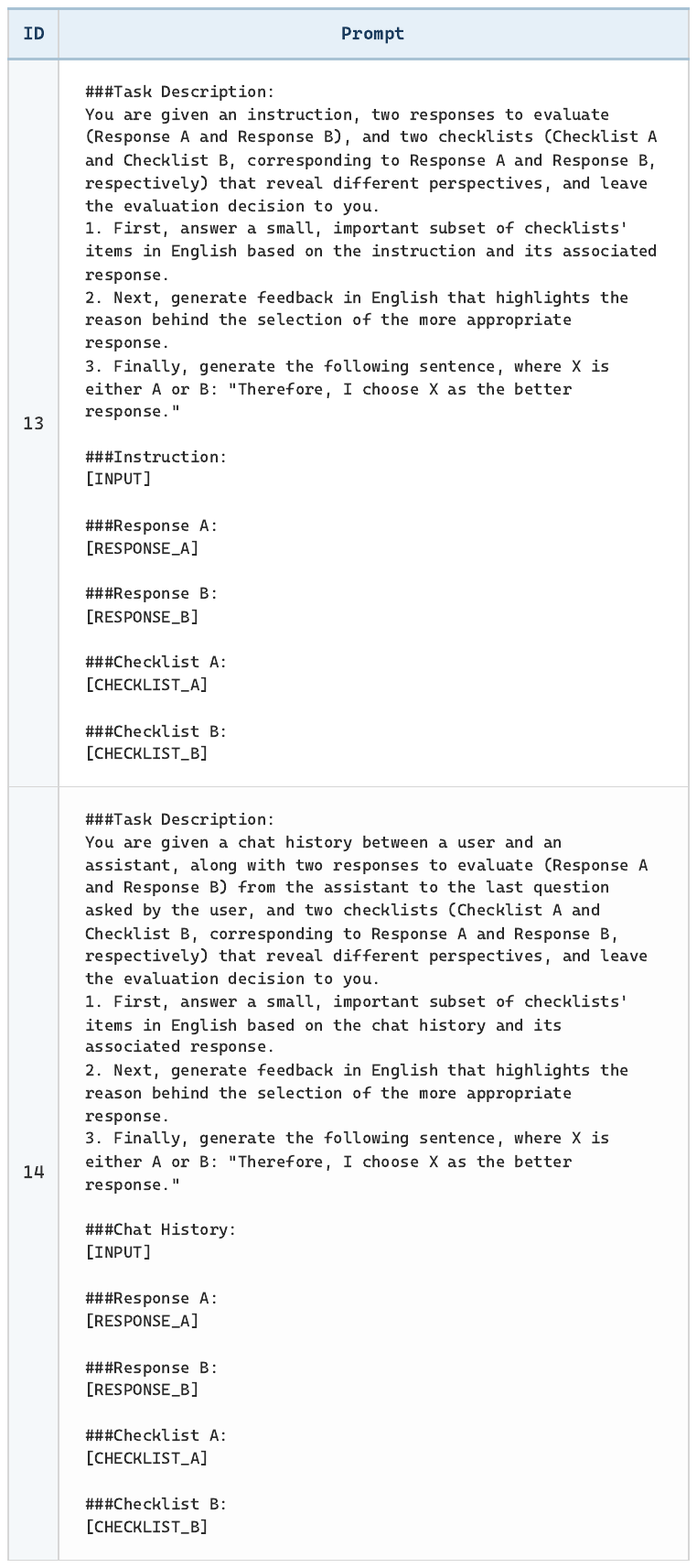}
	\caption{Judgment prompt for MM-Eval (Reasoning \& Chat) datasets.}
	\label{fig:judgepair}
\end{figure*}

We only use system prompts from this section, which are shown in \ref{fig:system}: one for the Liteval dataset and another for the MM-Eval datasets. Figure \ref{fig:judgelit} presents the prompt template for the Liteval dataset, while \ref{fig:judgepair} shows the prompts for the two MM-Eval datasets. In these prompts, the placeholders clearly indicate what should replace them. Importantly, to demonstrate the flexibility of our framework, we also use a scoring guide for the pointwise assessment to help our judge LLM perform a more accurate evaluation.

\end{document}